\documentclass[10pt,letterpaper]{article}
\usepackage[top=0.85in,left=2.75in,footskip=0.75in]{geometry}

\usepackage{amsmath,amssymb}

\usepackage{changepage}

\usepackage[utf8x]{inputenc}

\usepackage{textcomp,marvosym}

\usepackage{cite}

\usepackage{nameref,hyperref}

\usepackage[right]{lineno}

\usepackage{microtype}
\DisableLigatures[f]{encoding = *, family = * }

\usepackage[table]{xcolor}

\usepackage{array}

\usepackage{subfig}
\usepackage{multirow}
\usepackage{multicol}
\usepackage{todonotes}
\usepackage{soul,color}
\usepackage{algorithm}
\usepackage{algorithmic}
\newcommand{\INDSTATE}[1][1]{\STATE\hspace{#1\algorithmicindent}}
\newcolumntype{+}{!{\vrule width 2pt}}

\newlength\savedwidth

\raggedright
\usepackage[aboveskip=1pt,labelfont=bf,labelsep=period,justification=raggedright,singlelinecheck=off]{caption}

\makeatletter
\renewcommand{\@biblabel}[1]{\quad#1.}
\makeatother

\usepackage{lastpage,fancyhdr,graphicx}
\usepackage{epstopdf}
\fancyheadoffset[L]{2.25in}
\fancyfootoffset[L]{2.25in}
\begin{document}
\vspace*{0.2in}

\begin{flushleft}
{\Large
\textbf\newline{Interpretable PID Parameter Tuning for Control Engineering using General Dynamic Neural Networks: An Extensive Comparison}
}
\newline
\\
Johannes G\"unther\textsuperscript{1,2*},
Elias Reichensd\"orfer\textsuperscript{3},
Patrick M. Pilarski\textsuperscript{1,2},
Klaus Diepold\textsuperscript{3},
\\
\bigskip
\textbf{1} Department of Computing Science, University of Alberta, Edmonton, Alberta, Canada
\\
\textbf{2} Alberta Machine Intelligence Institute, Edmonton, Alberta, Canada
\\
\textbf{3} Department of Electrical and Computer Engineering, Technische Universit\"at M\"unchen, Munich, Bavaria, Germany
\\
\bigskip

%
%





* Corresponding author\\
E-mail: gunther@ualberta.ca (JG)

\end{flushleft}
\begin{abstract}

Modern automation systems largely rely on closed loop control, wherein a controller interacts with a controlled process via actions, based on observations. 
These systems are increasingly complex, yet most deployed controllers are linear Proportional-Integral-Derivative (PID) controllers. 
PID controllers perform well on linear and near-linear systems but their simplicity is at odds with the robustness required to reliably control complex processes.
Modern machine learning techniques offer a way to extend PID controllers beyond their linear control capabilities by using neural networks.
However, such an extension comes at the cost of losing stability guarantees and controller interpretability. 
In this paper, we examine the utility of extending PID controllers with recurrent neural networks—--namely, General Dynamic Neural Networks (GDNN); we show that GDNN (neural) PID controllers perform well on a range of complex control systems and highlight how they can be a scalable and interpretable option for modern control systems. 
To do so, we provide an extensive study using four benchmark systems that represent the most common control engineering benchmarks.
All control environments are evaluated with and without noise as well as with and without disturbances. 
The neural PID controller performs better than standard PID control in $15$ of $16$ tasks and better than model-based control in $13$ of $16$ tasks. 
As a second contribution, we address the lack of interpretability that prevents neural networks from being used in real-world control processes. 
We use bounded-input bounded-output stability analysis to evaluate the parameters suggested by the neural network, making them understandable for engineers. 
This combination of rigorous evaluation paired with better interpretability is an important step towards the acceptance of neural-network-based control approaches for real-world systems. 
It is furthermore an important step towards interpretable and safely applied artificial intelligence.
\end{abstract}

\section{Introduction}
Modern production engineering involves increasingly complex physical processes \cite{elmaraghy2012complexity}. 
The physical processes underlying cutting-edge production engineering cannot be appropriately expressed with simple models \cite{aastrom2000limitations}. 
New, more complex classical control methods are being developed, but their increased complexity is what makes them challenging for control engineers to design and apply\cite{klatt2009perspectives}. 
Limited by both the number of skilled control engineers and their cost, the production engineering industry has widely chosen to continue to use simple, understandable linear controllers. While these controllers are easy to set-up and adjust, they are not suitable for the complex non-linear behaviour of the processes they are expected to control. 
Using these simple controllers comes at a cost: it means processes will require close monitoring and human assistance whenever the system changes in an unforeseen way. 
In the face of increasingly complex systems, both control-engineer-designed methods and closely-monitored simple controllers fail to scale.

One way to bridge this gap between simple controllers and complex control systems is by applying modern machine learning techniques \cite{wuest2016machine}. Extending the capabilities of well-accepted and used controllers with machine learning yields a potential solution to the lack of scalability and adaptability in existing control approaches. In this paper, we will investigate the use of neural networks to adapt the parameters of a Proportional-Integral-Differential (PID) controller not only before deployment but online during ongoing control. This continuous adaptation to the process would allow for linear controllers to perform well for any control task, as the controller constantly linearizes its behaviour around the state the system is currently in. As our investigation shows, this results in superior control performance and disturbance rejection.

 The PID controller is one of the most widely used controllers \cite{ang_2005}. A PID controller calculates its control output $u$ based on the current error $e$, the error derivative $\frac{de}{dt}$ and the error integrated over time $\int_0^t e dt$. Each error measure is multiplied by its corresponding constant---$K_P$ with the error, $K_I$ with the integrated error, and $K_D$ with the error derivative---and then summed. A PID controller is tuned with respect to the system to be controlled by adjusting these three constants.
 While this controller is simple and well understood, its advantages come at the price of limited capabilities. PID controllers perform well only on linear systems or systems that are linearized. 
As PID controllers are usually adjusted before deployment, they neither handle disturbances nor varying system dynamics well enough to meet the needs of modern production engineering. 

Several prior studies have shown that neural networks can improve performance when used to tune the parameters of traditional PID controllers online. Simple feed-forward networks can adjust the PID parameters in multiple settings \cite{de2017modeling,de2018discrete}. More sophisticated neural networks like strictly recurrent, diagonal \cite{zhang_2006} and quasi-diagonal \cite{zhang_2010} recurrent networks have also been investigated. However, these specific design choices for the neural network's architecture result in very specific behaviour and learning \cite{bergstra2013making} and thus are unlikely to generalize well over different control environments. To evaluate whether the use of neural networks with all possible connections is applicable to control engineering tasks in general, a broad range of different control environments is required. In this paper, we therefore implement and evaluate general dynamic neural networks, thus extending the current state of the art.

Previous works have demonstrated the applicability of neural-network-based parameter tuning, but only for one well-defined control task at a time, e.g. pendulums \cite{de2018discrete}, two-tank systems \cite{pirabakaran2002pid} or magnetic systems \cite{de2017modeling}. As no single approach has been applied successfully to multiple representative control engineering problems in the literature, the effectiveness of the suggested approaches is hard to estimate. For this reason, we investigate our approach on four different systems that represent the most common challenges in control engineering, resulting in an extensive comparison. By comparing the control performance on all four different tasks, we extend the current literature which focused on one problem at a time.
Additionally, our simulations are closer to real-world conditions than those used in previous work, as they include disturbances and noise.

Despite their potential to improve control performance, neural networks are not widely used for parameter tuning in real-world control systems. One major barrier prevents their adoption: due to the lack of interpretability, the suggested PID parameters cannot be evaluated for their appropriateness. As a result, the control loop itself loses one of its most important properties---its stability guarantees. Previous papers that utilized neural networks for PID parameter tuning did not address the effect on input-output stability, thus ignoring a key concern in control engineering. The domain of control engineering demands interpretability to ensure system safety \cite{leveson2003systems}. As neural networks introduce a black box to the system \cite{yang2019enhancing}, their use for PID parameter tuning is not yet widely accepted. Real-world use requires the ability to check and reason about the parameter choices a neural network outputs when granted access to a PID controller's parameters. We suggest that the use of neural networks will in fact enable designers to achieve better performance without significantly increasing the design complexity or diminishing design interpretability when deploying a controller---if combined with traditional control engineering tools like stability analysis. 

To summarize our contributions:
as a first contribution of this paper, we advance the use of specifically tailored neural network architectures by investigating the use of General Dynamic Neural Networks (GDNNs) for online parameter tuning in PID controllers. These neural PID controllers are evaluated on four different closed loop control engineering tasks. Each task represents a different, common challenge in control engineering, namely non-linear behaviour, unstable equilibrium, dead time, and chaotic behaviour. 
These applications have furthermore been used for evaluation of control at several machine learning venues, e.g. \cite{yu2014efficient,berkenkamp2017safe,parmas2018pipps}. We compare the performance of our neural approach with a standard PID controller, which acts as a baseline, and with a system-appropriate model-based controller, which should provide the best possible performance. 
To evaluate the robustness of our approach, each comparison is performed with and without significant sensor noise and with and without disturbance. All controllers are evaluated quantitatively for all scenarios, making this study unique in its comprehensiveness when compared to prior work. 

As a second contribution, we demonstrate how input-output stability analysis---a classic analysis that is well known in control engineering---can be used in a novel, online way to explain the effects of parameter tuning. This analysis is done for the control system, exhibiting chaotic behaviour as an example of a challenging control task.
This assessment provides an explanation of when the system is stable with respect to the PID parameters, making the neural network outputs understandable to control engineers. While input-output stability is an important aspect of every closed-loop control approach, it is widely ignored in the existing literature that suggests the use of neural networks for PID parameter tuning.
This paper therefore addresses key attributes of safe and interpretable artificial intelligence.

\section{General Dynamic Neural Network for PID Parameter Tuning}
\begin{figure*}[!t]
    \centering
    \subfloat[Closed loop control]{\resizebox{0.45\linewidth}{!}{\includegraphics[width=\columnwidth]{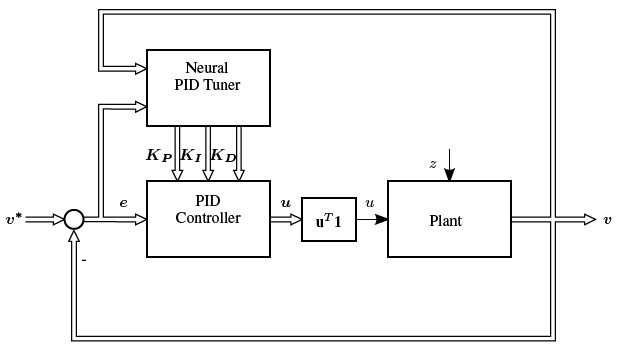}}}\hfil
    \subfloat[Example neural network]{\resizebox{0.4\linewidth}{!}{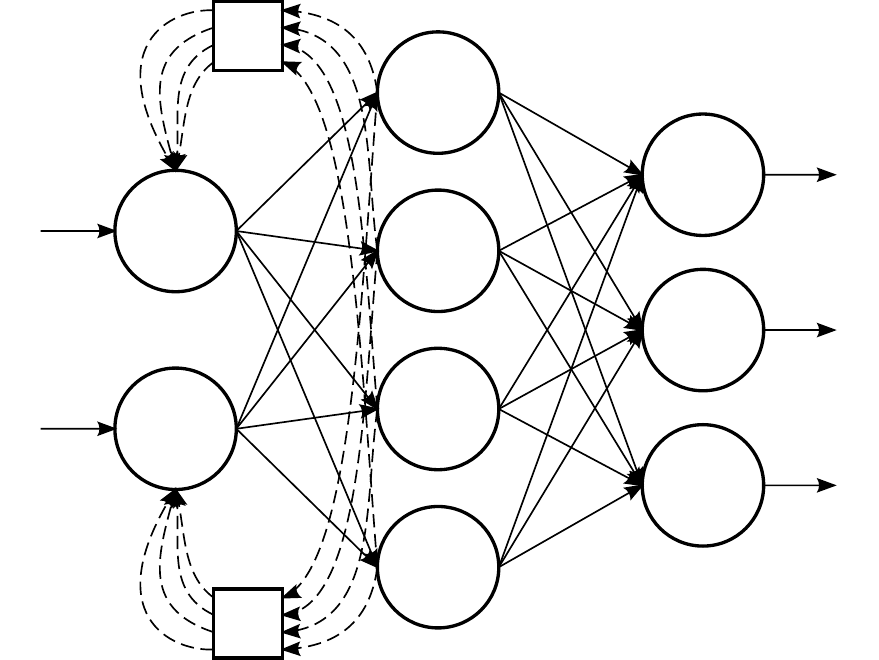}}\
    \caption{\textbf{Neural PID architecture and example neural network.}
    A neural network is integrated into standard closed loop control (a). The neural network receives the system output and the error as input and outputs the three PID parameters $K_P$, $K_I$, and $K_D$. 
    The double-lined arrows indicate that the associated variable could be a vector, while the single-lined arrows indicate scalar variables. An example neural network (b) shows one possible set of connections. All networks have $9$ neurons in three layers. The output of each input neuron is fed back into the input layer with a delay, denoted as $g^{-1}$, of one time step.} \label{fig:closed_loop_and_network}
\end{figure*}
The aim of this work is to extend the classic PID controller framework, composed of the PID controller and the plant, with as few changes to the physical setup of the control system as possible. 
Maintaining the core structure of the classic PID control framework allows for easy adoption to existing industrial applications. This structure is preserved by restricting the neural network's inputs to signals which are already available in the closed loop control setting. The closed loop control setting is depicted in Figure \ref{fig:closed_loop_and_network}(a).
The neural network's inputs are the control system's output $v$ and the control error $e$, while the outputs are the three PID parameters to tune, i.e. $K_P$, $K_I$, and $K_D$. 
At each time step, the neural network computes the PID parameters based on the observations and passes them into the PID controller. 
The PID controller then uses these parameters to compute the control output $u$.

To ensure fast online computation with limited hardware, the neural networks implemented in this investigation are restricted to one hidden layer with four neurons. 
Furthermore, for control systems that can be described by simple differential equations increasing the number of neurons would lead to overfitting. The network architectures differ only in additional recurrent or feedback connections. An example can be seen in Figure \ref{fig:closed_loop_and_network}(b). The figure shows a neural network, where the output of each neuron in the first layer is fed back as part of the first layer's input with a delay, denoted as $g^{-1}$, of one time step. For all neurons, the activation function \texttt{tanh} is chosen, following the rationale of \cite{kalman1992tanh}.

The standard approach for training neural networks is backpropagation \cite{rumelhart1995backpropagation}.
Most deep learning approaches adjusts the neural network's weights by end-to-end optimization \cite{glasmachers2017limits}. This optimization involves formulating a loss function that describes the difference between the neural network's outputs and the ideal outputs. From this loss, a gradient with respect to the weights is computed and propagated through the network. At each neuron, the weights are then adjusted to minimize the loss. 
However, in the present framework, Figure \ref{fig:closed_loop_and_network}(a), backpropagation cannot be naively applied. In this framework, the neural network's outputs are the PID parameters $K_P$, $K_I$, and $K_D$. Using standard backpropagation would therefore require knowing the ideal PID parameters at any given time.

A way to train the neural network without the ideal outputs is to numerically approximate backpropagation. In this work, we chose a numerical Levenberg-Marquard algorithm \cite{marquardt_1963} to minimize the squared control error.
At each time step, the Jacobian matrix, shown in Equation \ref{eq:jacobian}, of the neural network weights is numerically approximated using finite differences with respect to small changes in each weight. Each weight is then adapted to decrease the control error.

The Jacobian matrix of a general dynamic neural network with $p$ input neurons, $q$ dynamical system (plant) outputs and k weighted connections was calculated as
\begin{equation}
\label{eq:jacobian}
J_{\textbf{w}}\big(\textbf{v}(\textbf{x}; \textbf{w})\big) = 
\begin{bmatrix}
\frac{\partial}{\partial w_0}v_0(x_0; \textbf{w}) & \frac{\partial}{\partial w_1}v_0(x_0; \textbf{w}) & \cdots & \frac{\partial}{\partial w_k}v_0(x_0; \textbf{w}) \\
\frac{\partial}{\partial w_0}v_0(x_1; \textbf{w}) & \frac{\partial}{\partial w_1}v_0(x_1; \textbf{w}) & \cdots & \frac{\partial}{\partial w_k}v_0(x_1; \textbf{w}) \\
\vdots & \vdots & \ddots & \vdots \\
\frac{\partial}{\partial w_0}v_0(x_p; \textbf{w}) & \frac{\partial}{\partial w_1}v_0(x_p; \textbf{w}) & \cdots & \frac{\partial}{\partial w_k}v_0(x_p; \textbf{w}) \\
\frac{\partial}{\partial w_0}v_1(x_0; \textbf{w}) & \frac{\partial}{\partial w_1}v_1(x_0; \textbf{w}) & \cdots & \frac{\partial}{\partial w_k}v_1(x_0; \textbf{w}) \\
\frac{\partial}{\partial w_0}v_1(x_1; \textbf{w}) & \frac{\partial}{\partial w_1}v_1(x_1; \textbf{w}) & \cdots & \frac{\partial}{\partial w_k}v_1(x_1; \textbf{w}) \\
\vdots & \vdots & \ddots & \vdots \\
\frac{\partial}{\partial w_0}v_1(x_p; \textbf{w}) & \frac{\partial}{\partial w_1}v_1(x_p; \textbf{w}) & \cdots & \frac{\partial}{\partial w_k}v_1(x_p; \textbf{w}) \\
\vdots & \vdots & \ddots & \vdots \\
\frac{\partial}{\partial w_0}v_q(x_0; \textbf{w}) & \frac{\partial}{\partial w_1}v_q(x_0; \textbf{w}) & \cdots & \frac{\partial}{\partial w_k}v_q(x_0; \textbf{w}) \\
\frac{\partial}{\partial w_0}v_q(x_1; \textbf{w}) & \frac{\partial}{\partial w_1}v_q(x_1; \textbf{w}) & \cdots & \frac{\partial}{\partial w_k}v_q(x_1; \textbf{w}) \\
\vdots & \vdots & \ddots & \vdots \\
\frac{\partial}{\partial w_0}v_q(x_p; \textbf{w}) & \frac{\partial}{\partial w_1}v_q(x_p; \textbf{w}) & \cdots & \frac{\partial}{\partial w_k}v_q(x_p; \textbf{w}) 
\end{bmatrix} \in \mathbb{R}^{pq \times k}
\end{equation}
, where $\textbf{x} \in \mathbb{R}^p$ is the neural network input vector, $\textbf{w} \in \mathbb{R}^k$ is the vector of weights which describes the network topology and $\textbf{v}(\textbf{x}; \textbf{w}) \in \mathbb{R}^{q}$ are the outputs of the dynamical system (plant). According to \cite{de_2007}, it is sufficient to calculate the partial derivatives of the systems output instead of the error function.

As the analytic calculation would result in extensive computations, it is numerical approximated using a difference equation, rather than a differential one as solution of the equation
\begin{equation}
\label{eq:jacobiannum}
\hat{J}_{\textbf{w}}\big(\textbf{v}(\textbf{x}; \textbf{w})\big) = 
\begin{bmatrix}
\textbf{j}_0 \\
\textbf{j}_1 \\
\vdots \\
\textbf{j}_{k}
\end{bmatrix}^{T} \, , \,
\textbf{j}_i = \frac{\textbf{v}(\textbf{x}; \textbf{w}) - \textbf{v}(\textbf{x}; \textbf{w} - \eta\varepsilon(\textbf{w}_i))}{\varepsilon(\textbf{w}_i)} \, , \, \eta_p = \left\{\begin{array}{ll} 1, & p = i \\
           0, & p \neq i \end{array}\right. ,
\end{equation}
where $\hat{J}_{w}$ is the approximated Jacobian matrix, $\textbf{j}_i$ is the $i$-th column of $\hat{J}_{w}$, $\eta$ is the step size and $\epsilon$ is the machine precision. 
$\epsilon$ has to be calculated for each pass, using the equation
\begin{equation}
\label{eq:optimaleps}
\epsilon(\textbf{w}_i) = \max\big(1, |\textbf{w}_i|\big)\sqrt{\epsilon_{\text{min}}}, 
\end{equation}
where $\textbf{w}$ is a vector, containing the weights, and $\epsilon_{\text{min}}$ is the implementation data type, double precision in this implementation. The complete calculation of the Jacobian matrix can be found in Algorithm \ref{jacobian_alg}.

\begin{algorithm}[t!]
\caption{\textbf{Numerical approximation for Jacobian matrix.}}
\begin{algorithmic}[1]
\STATE \textbf{Input:} Dynamical system (plant) output $\textbf{v}(\textbf{x}; \textbf{w})$, neural network inputs $\textbf{x}$, weights $W$ \\
\STATE \textbf{Output:} Estimate of Jacobian matrix $\hat{J}_{W}$\\
\STATE \textbf{foreach:}Weight $\textbf{w}_i$ \textbf{do} 	\\
  \INDSTATE[1] $\epsilon \leftarrow \max\big(1, |\textbf{w}_i|\big)\sqrt{\epsilon_{\text{min}}}$ \hspace{40mm} // Calculate $\epsilon$\\	
  \INDSTATE[1] \textbf{For} $j = 1$ to $g$ \textbf{do}   \\
  \INDSTATE[2] $\textbf{v}_{\text{tmp}, 1} \leftarrow \textbf{v}(\textbf{x}; \textbf{w})$, $\textbf{v}_{\text{tmp}, 2} \leftarrow \textbf{v}(\textbf{x}; \textbf{w} - \eta\varepsilon(\textbf{w}_i))$ \hspace{2mm} // Compute difference of $\textbf{v}$\\
    \INDSTATE[2] \textbf{For} $o = 1$ to $q$ \textbf{do}   \\
        \INDSTATE[3] $\hat{J}_{W(jm + o, i)} \leftarrow \frac{\textbf{v}_{\text{tmp}, 1} - \textbf{v}_{\text{tmp}, 2}}{\epsilon}$ \hspace{27mm} // Backward difference\\
\STATE \textbf{Return:} $\hat{J}_{W}$  \\
    
\end{algorithmic}
\label{jacobian_alg}
\end{algorithm}

    

In the industrial control setting we consider in this paper, it is important that finding an appropriate network architecture, i.e. the connections and delays between neurons, does neither need sophisticated engineering nor significant time. Any approach that makes the set up too complicated would defeat the purpose of extending the existing PID framework. 
We therefore chose to create ten neural network architectures for each control challenge by randomly adding feedback connections and then chose the neural network that performed best. This process of finding an architecture can be made more efficient by using search algorithms \cite{sandner2017automated}. 
For each tested neural network, the weights were initialized randomly with a mean of zero and a standard deviation of one.

Training data was collected from the differential equations, that describe each benchmark system, without any additional noise or disturbances. We followed an approach from classical control engineering by exciting the system with input signals of different lengths and amplitudes---called Amplitude Modulated Probabilistic Random Binary Signals (APRBS) \cite{deflorian2011design}---to collect data samples that sufficiently describe the dynamics of the system \cite{guenther_2018}. This approach is an extension of using a Dirac impulse for system identification \cite{aastrom1971system}. We collected $35,000$ data samples for each benchmark system, and split them into training, validation, and test sets with a ratio of $15\%$, $30\%$, and $55\%$, respectively.

\section{Experiments}
To evaluate the performance of the neural PID controller, we use four typical control problems. Each system offers a different control challenge. Each individual system is controlled with and without noise. The noise is Gaussian white noise with a signal to noise ratio (SNR) of $20$dB and corresponds to noisy measurements from sensors and is added to the system output.

To further evaluate the robustness of the controller, we inflict upon each system a suitable disturbance. The term disturbance refers to an unwanted and unexpected system input that will result in an increase of the system error. Disturbances can occur due to external influences or due to failure in the system and they are individual for each kind of system. For each system, the disturbance is increased in size from zero until only one control approach is still able to stabilize the system. Disturbances of this magnitude are then used in the experiments. These disturbances are not included in the training data for the neural PID controller. All experiments are run over $50$ independent runs to ensure statistical relevance. The system differential equations are solved using the Dormand-Prince solver. To simulate real-world conditions of a discrete sample time, the controller output can be adapted every $0.01$s. This intervention time is chosen to represent the limitations of real-world actuators, which cannot adjust their values on an arbitrarily small timescale. The solver is run iteratively for $0.01$s, using the result of the former step as starting conditions. During each $0.01$s intervention time window, the controller output $u$ is kept constant.

\begin{figure}[!t]
    \centering
    {\resizebox{.5\linewidth}{!}{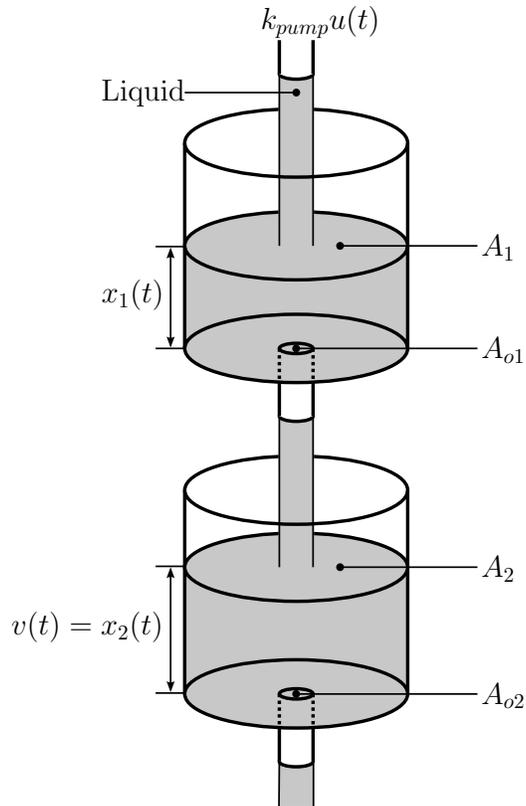}}
    \caption{\textbf{Two-Tank system.} This nonlinear system is widely used to study nonlinear behaviour in control engineering systems. The controller can adjust the amount of water being put into the first tank. The goal is to keep the water level in the lower tank ($v(t)=x_2(t)$ at the setpoint.} 
    \label{fig:two_tank}
\end{figure}

The training is done on an Intel Core i5-4570 with a $3.2$ GHz clock rate, $6$ MB of shared L$3$ cache, $32$ GB DDR3 RAM. Once learned, the neural networks run on a Raspberry Pi $3$ Model A+.
\textbf{Two-Tank System:} The first system is a nonlinear two-tank system, as seen in Figure \ref{fig:two_tank} and is described by the differential equations in \cite{pan_2005}. The controller has access to a pump, regulating the input, while the measured output is the water level in the second tank. This system is a standard benchmark system in control theory. It corresponds to various industrial processes, e.g. bio-reactors, filtration, and nuclear power plants. There exist a number of control approaches for this system, including direct control via neural networks \cite{majstorovic_2008}, adaptive output feedback \cite{Takagi_2014}, and backstepping \cite{pan_2005}, which will be used as comparison.

To evaluate the robustness of the compared control approaches, the two-tank system is disturbed continuously between $t = 20$s and $t = 40$s. As a disturbance, the controller output is set to zero, which would correspond to a drain of the water supply. The water levels in the tank are therefore independent from the control inputs for $20$s. The voltage for the pump, $u$, was limited to the range $[-500\text{V}, 500\text{V}]$ to simulate a pump appropriate to the tank dimensions.

\begin{figure}[!ht]
    \centering
    {\resizebox{.5\linewidth}{!}{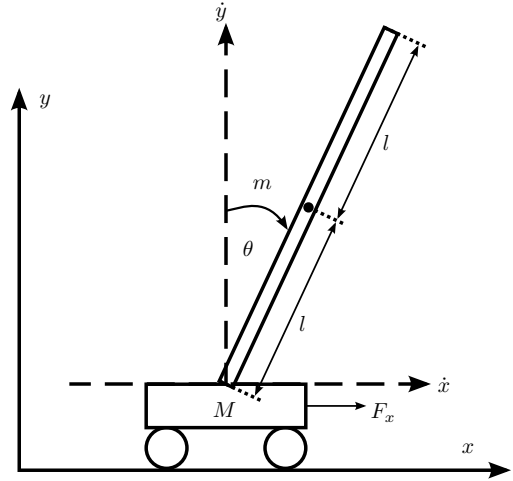}}
    \caption{\textbf{Inverted Pendulum on a Cart.} This system is a nonlinear system with an unstable equilibrium. The control task is to move the cart to a predefined position, while keeping the pole up.}
    \label{fig:inverted_pendulum}
\end{figure}

\textbf{Inverted Pendulum on a Cart:} The second system is a nonlinear inverted pendulum on a cart, as shown in Figure \ref{fig:inverted_pendulum}. The system is described by the differential equations in \cite{wang_2011}. The control task is to stabilize the inverted pendulum at its unstable equilibrium by applying a force on the cart. The cart's movement is restricted to $0.5$m in either direction. This system is a widespread benchmark system in control theory due to its nonlinearity and unstable equilibrium \cite{wang_2011,kim_2017}. Practical applications for inverted pendulums include rocket control during initial stages of flight or keeping a walking robot in an upright position. For comparison, a linear–quadratic regulator (LQR) \cite{jaleel_2013,li_2012} and a double PID controller \cite{li_2012} are used.

The system is disturbed by a force of $8.5$N to the pendulum at time $t=10$s. This disturbance can be interpreted as a strong and unexpected wind condition during the launch of a rocket. The controller ouptut $u$ was bounded within $[-50\text{N}, 50\text{N}]$, which corresponds to a typical actuator of that size.

\begin{figure}[!ht]
    \centering
    {\resizebox{.75\linewidth}{!}{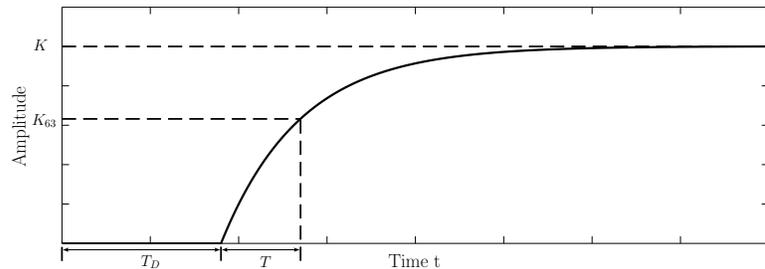}}
    \caption{\textbf{System with Non-negligible Time Delay.} The Figure shows the systems response (including the time delay $T_D$) to an input.}
    \label{fig:time_delay}
\end{figure}

\textbf{System with Non-negligible Time Delay:} The third system is a first order linear time invariant (LTI) system with a non-negligible time delay. Time delay is a problem in control theory that is often forgotten while designing controllers \cite{wang_2013}. Time delays can result in decreased performance and system instability. The benchmarks for this system are a PID controller \cite{tavakoli_2003} and a smith-predictor \cite{stojic_2001}. Figure \ref{fig:time_delay} demonstrates the delayed system response for an input.

This system is disturbed by a (dimensionless) disturbance of $-5$ between $t = 50$s and $t = 75$s continuously. Such a disturbance can be thought of as a temporary blockage in a fluid transport system. The exact system specifications can be found in \cite{tavakoli_2003}. The controller output $u$ was bounded between the range of $[-10, 10]$.

\begin{figure}[!t]
    \centering
    {\resizebox{.75\linewidth}{!}{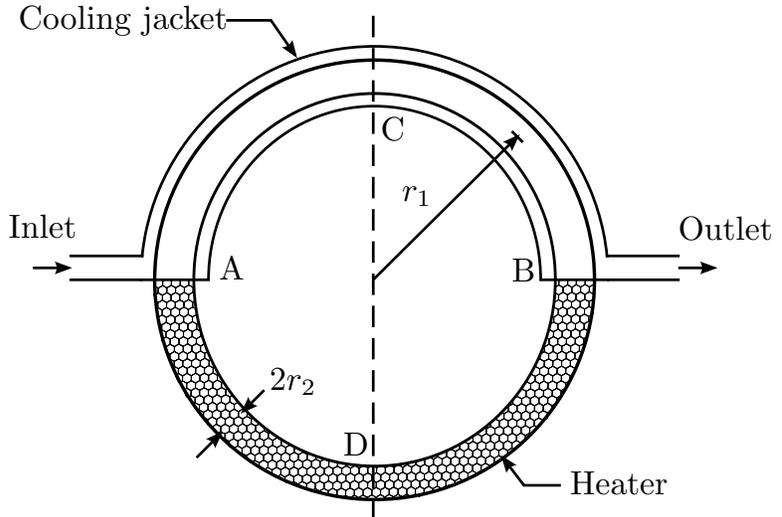}}
    \caption{\textbf{Chaotic Thermal Convection Loop.} This system is an example for chaotic behaviour. The control task is to maintain a constant flow in the inner torus - the flow is measured at the points A and B. Half of the torus that contains the fluid is surrounded by a heating element, the other half is surrounded by a cooling jacket. The control variable is the heating power, applied on the lower half of the torus.}
    \label{fig:fluid_system}
\end{figure}

\textbf{Chaotic Thermal Convection Loop:} The fourth system is a chaotic thermal convection loop, as shown in \ref{fig:fluid_system}. Its dynamics are described by the equations 
\begin{equation} 
\label{eq:fluidsystem}
\begin{split}
\dot{x_1}(t) & = p(x_2 - x_1) \, , \\
\dot{x_2}(t) & = x_1 - x_2 - x_3(x_1 + \beta) , \\
\dot{x_3}(t) & = x_1x_2 + \beta(x_1 + x_2) - x_3 - u, \\
\end{split}
\end{equation}
with $p= 10$ and $\beta = 6$ as appropriate constants \cite{creveling1975stability}. $x_1$ is a measure for how far the current flow velocity differs from the steady point of the system---if $x_1$ is zero, the system is in its steady point. $x_2$ and $x_3$ are measures for the difference in temperature between the points A and B, as well as C and D in Figure \ref{fig:fluid_system} respectively. $u$ is a the power, applied to the heater and the control variable. 

Chaotic behavior may lead to vibrations, oscillations and failure in systems and is therefore an important aspect of control theory. As chaotic behavior is unpredictable, mathematical models are only sufficient to a certain point, hence closed loop control is a desirable approach \cite{wang_1992}. Usual control approaches for the chaotic thermal convection loop are nonlinear feedback controllers \cite{boskovic_2000} and backstepping \cite{vazquez_2006,vazquez_2010}.

To evaluate the robustness of the applied control approaches, the system is disturbed with a force of $100$W continuously between $t = 5$s and $t = 5.5$s. This perturbation can be interpreted as a temporary change in the cooling water temperature. To simulate real actuators with a limited capacity, the controller output $u$ is limited between $[-100\text{W}, 100\text{W}]$ to simulate an appropriate heating element.

\section{Results and Discussion}
\begin{table*} [!t]
\centering
\caption{\textbf{Control results for the four benchmark systems.}}
\begin{tabular}{ l l l l l l }
	\hline
  \multicolumn{2}{l}{\textbf{Control Benchmark}} & \multicolumn{4}{c}{\textbf{RMSE on test data over 50 runs}} \\
	\hline
	\hline
	\multicolumn{2}{l}{\textbf{Disturbance}} & \multicolumn{1}{c}{-} & \multicolumn{1}{c}{-} & \multicolumn{1}{c}{\checkmark} & \multicolumn{1}{c}{\checkmark} \\
	\hline
	\multicolumn{2}{l}{\textbf{SNR}} & \multicolumn{1}{c}{-} & \multicolumn{1}{c}{20 dB} & \multicolumn{1}{c}{-} & \multicolumn{1}{c}{20 dB} \\
	\hline
	\hline
	\multicolumn{2}{l}{Two-tank system} & \multicolumn{4}{c}{} \\
	\hline
	\multirow{3}{*}{\rotatebox{90}{Mean}} & Neural PID & $\boldsymbol{0.74}$ & $\boldsymbol{0.86}$ & $\boldsymbol{0.84}$ & $\boldsymbol{1.00}$ \\
   & Standard PID & $0.95$ & $0.99$ & $1.00$ & $1.10$ \\
   & Backstepping & $1.10$ & $1.10$ & $1.20$ & $1.10$ \\
	\hline
	\multirow{3}{*}{\rotatebox{90}{Variance}} & Neural PID & $0.0036$ & $0.0040$ & $0.0056$ & $0.0460$ \\
	& Standard PID & $0.0025$ & $0.0023$ & $0.0052$ & $0.0063$ \\
	& Backstepping & $0.0016$ & $0.0018$ & $0.0036$ & $0.0024$ \\
	\vspace{-8pt}\\
	\hline
	\hline
	\multicolumn{2}{l}{Inverted pendulum} & \multicolumn{4}{c}{} \\
	\hline
	\multirow{3}{*}{\rotatebox{90}{Mean}}  & Neural PID & $\boldsymbol{0.034}$ & $\boldsymbol{0.02}$ & $\boldsymbol{0.09}$ & $\boldsymbol{0.27}$ \\
	& Standard PID & $\boldsymbol{0.035}$ & $0.04$ & $140$ & $140$ \\
	& LQ regulator & $0.05$ & $0.05$ & $140$ & $140$ \\
	\hline
	\multirow{3}{*}{\rotatebox{90}{Variance}}  & Neural PID & $0.0004$ & $0.0003$ & $0.0490$ & $0.0007$ \\
	& Standard PID & $0$ & $1.1\cdot 10^{-7}$ & $0$ & $0.0023$ \\
	& LQ regulator & $0$ & $3.3\cdot 10^{-9}$ & $0$ & $0.0022$ \\
	\vspace{-8pt}\\
	\hline
	\hline
	\multicolumn{2}{l}{LTI system with input delay} & \multicolumn{4}{c}{} \\
	\hline
	\multirow{3}{*}{\rotatebox{90}{Mean}}  & Neural PID & $\boldsymbol{0.13}$ & $\boldsymbol{0.15}$ & $0.26$ & $0.28$ \\
	& Standard PID & $0.22$ & $0.23$ & $0.36$ & $0.38$ \\
	& Smith predictor & $0.18$ & $0.19$ & $\boldsymbol{0.19}$ & $\boldsymbol{0.20}$ \\
	\hline
	\multirow{3}{*}{\rotatebox{90}{Variance}}  & Neural PID & $0.0006$ & $0.0005$ & $0.0007$ & $0.0003$ \\
	& Standard PID & $0.0007$ & $0.0006$ & $0.0004$ & $0.0003$ \\
	& Smith predictor & $0.0006$ & $0.0005$ & $0.0005$ & $0.0003$ \\
	\vspace{-8pt}\\
	\hline
	\hline
	\multicolumn{2}{l}{Chaotic thermal convection loop} & \multicolumn{4}{c}{} \\
	\hline
	\multirow{3}{*}{\rotatebox{90}{Mean}}  & Neural PID & $\boldsymbol{0.23}$ & $0.90$ & $\boldsymbol{1.90}$ & $\boldsymbol{1.70}$ \\
	& Standard PD & $0.24$ & $1.60$ & $13.00$ & $4.20$ \\
	& Backstepping & $0.26$ & $\boldsymbol{0.89}$ & $9.80$ & $9.80$ \\
	\hline
	\multirow{3}{*}{\rotatebox{90}{Variance}}  & Neural PID & $1.2\cdot 10^{-5}$ & $0.0003$ & $0.16$ & $0.22$ \\
	& Standard PD & $0$ & $2.00$ & $0$ & $1.40$ \\
	& Backstepping & $0$ & $0.0001$ & $0$ & $1.4\cdot 10^{-5}$ \\
	\vspace{-8pt}\\
	\hline
	\label{tab:ctrl_results}
\end{tabular}
\end{table*}
The results for all experiments can be found in Table \ref{tab:ctrl_results}. For each system the mean and variance from $50$ independent runs are shown for all controllers in all tested scenarios. The best control approach is highlighted in bold. For all values, a two-sample t-test was performed and a control approach is only considered to be superior for $p < 0.05$. From Table \ref{tab:ctrl_results}, it can be seen that the neural PID controller performs best in $12$ scenarios, pairs with the standard PID controller in one and performs second best in two scenarios. To compare control approaches, there are common measurements that are used in control engineering, e.g. rise time, overshoot, settling time \cite{ang_2005}. However, these measurements all address the error between the setpoint and the actual systems output with different emphases. 
We therefore used the root-mean-squared-error (RMSE) between the setpoint and the system output to summarize these error measures in a single number without losing information.
\begin{figure*}[!t]
    \centering
    \resizebox{\linewidth}{!}{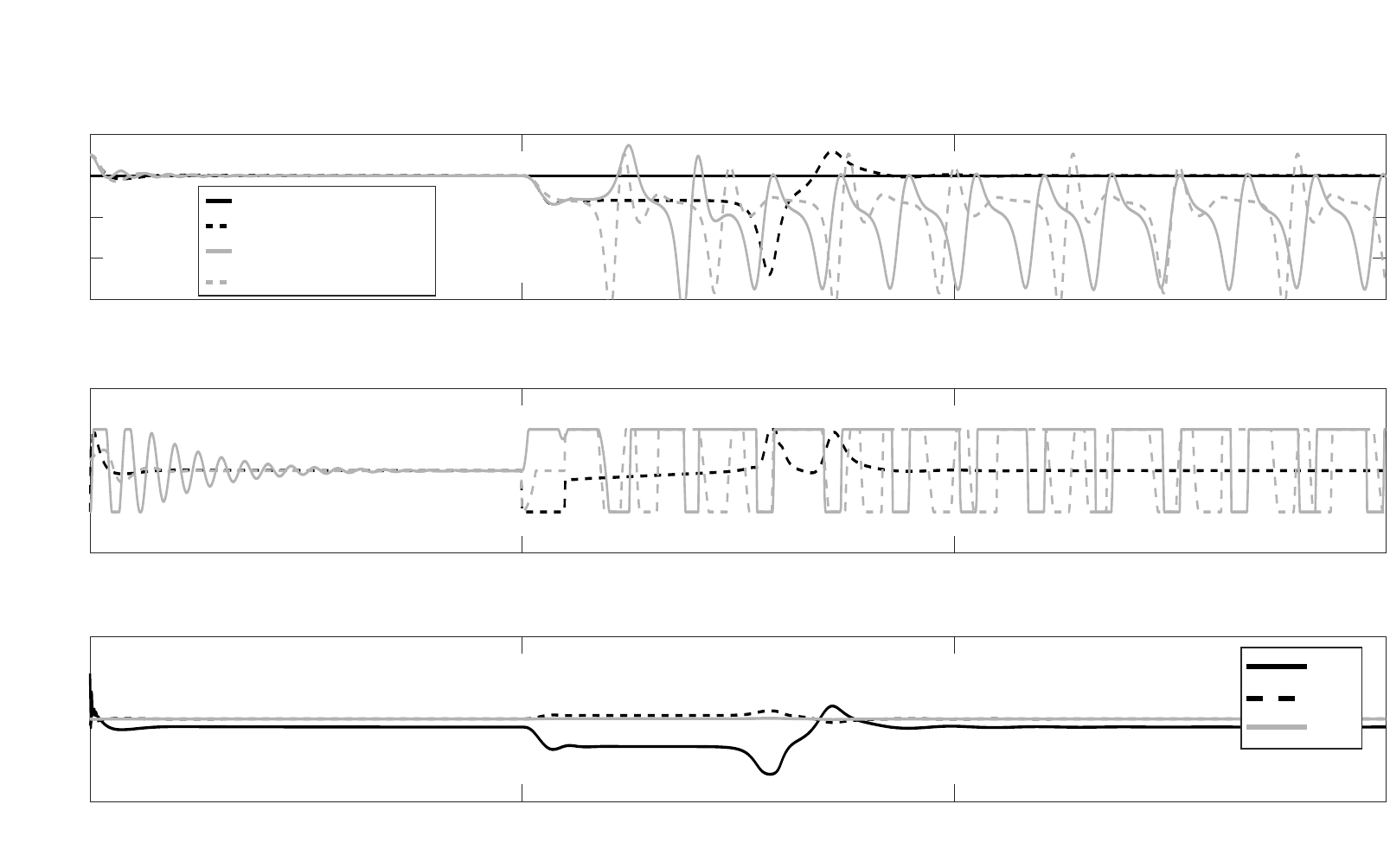}
    \caption{\textbf{Control performance for the disturbed chaotic thermal convection loop.} Subfigure (a) shows the setpoint ($x_1 = 0$, which corresponds to a steady flow) and the system output for all controllers. Subfigure (b) shows the controller outputs for all three controllers. In subfigure (c), the PID parameters, applied by the neural network are shown.}
    \label{fig_control_results}
\end{figure*}

\textbf{Two-Tank System:} For the two-tank system, a backstepping controller is chosen for a comparison, as this approach takes the nonlinear behavior of the system into account and has demonstrated to be well suited for this system \cite{pan_2005}. The PID controller is parameterized with the constants $K_P=3.65, K_D=-2$ and $K_I=0.4$. All controllers are able to control the system, while the neural PID controller exhibited the smallest error for all scenarios. However, the advantage the neural PID controller yields is relatively small, as can be seen in Table \ref{tab:ctrl_results}. As this system is the easiest to control, it can be expected that the standard PID controller and backstepping perform on a similar scale.

\textbf{Inverted Pendulum:} The inverted pendulum on a cart is controlled by a standard PID controller stack \cite{wang_2011}, and a LQR \cite{li_2012} for benchmarking. The PID controller, responsible for the position, has the values are $K_P=-2.4, K_D=-0.75$ and $K_I=-1$ and the controller for the angle is set to $K_P=25, K_D=3$ and $K_I=15$ \cite{wang_2011}.
All three approaches are able to initially stabilize the system. For the scenario without disturbance, the neural PID controller performs equally to the standard PID controller when there is no noise present. For the scenario with only noise, the neural PID controller is the best control approach. Furthermore, only the neural PID controller is able to stabilize the disturbed system, resulting in a substantially lower error as can be seen in Table \ref{tab:ctrl_results}.

\textbf{System with Non-negligible Time Delay:} For the system with non-negligible time delay, the standard PID controller is set to $K_P=1.5, K_D=-0.1$ and $K_I=0.7$. The neural PID controller is superior in all scenarios, when compared to the standard PID controller. When compared to the Smith predictor \cite{smith_1957}, the neural PID controller performs better only for the scenarios without noise. However, as the Smith predictor has knowledge about the exact time delay, it has a significant advantage over the neural PID controller.

\textbf{Chaotic Thermal Convection Loop:}
The PID parameters yielding the lowest error for the chaotic thermal convection loop are $K_P = 25.3$, $K_D = 8.9$ and $K_I=0$---the controller is therefore a PD controller.
The system is initialized outside of its inherently stable region (region of attraction) with the initial conditions $x_1 = x_2 = x_3 = 5$. Without control, the system will therefore not converge to the desired steady state $x_1 = x_2 = x_3 = 0$.

All controllers are capable of stabilizing the system, as can be seen in Figure \ref{fig_control_results}(a). The backstepping approach has the least overshoot but takes a long time to reach the steady state. The standard PID controller is more aggressive, resulting in a higher overshoot but still a smaller error. 
The neural PID controller performs best as it finds a good balance between settling time and overshoot.

Out of four scenarios, the neural PID controller demonstrates superior control performance. Only in the scenario with noise but without disturbance does backstepping perform slightly better ($0.89$ vs $0.9$). This can be explained by backstepping being designed using the differential equations. It therefore knows the underlying systems dynamics and is less influenced by the sensor noise.

Between the time $t=5$s and $t=5.5$s, the control output is set to $u = -100$W to simulate the disturbance described earlier. The standard PD controller becomes meta stable and its controller output iterates between the maximum value of $100$W and the minimum value of $-100$W. Although backstepping is proven to be globally, asymptotically stable in the Lyapunov sense \cite{mascolo1999controlling}, it also becomes meta stable. This can be explained by the real world conditions. As the controller can change its control output only every $0.01$s the backstepping approach fails, resulting in switching inputs between the maximum value and the minimum value, as seen in Figure \ref{fig_control_results}(b). Both controllers (PID and backstepping) use excessive amounts of energy without being able to stabilize the system.

The neural PID controller is able to stabilize the system after the disturbance. Figure \ref{fig_control_results}(c) shows how the neural network changes the PID parameters in response to the system output. When $x_1$ is far from the setpoint, the $K_P$ parameter has a high absolute value to force the system towards its steady state. To further increase the controller output at $t = 7.9$s, where the system reaches its furthest distance from the set point $K_I$ is increased. After the system reaches its steady state again, all PID parameters are adjusted back to their stationary value to ensure asymptotic performance. The neural PID controller furthermore uses significantly less energy to control the system.
\begin{figure*} [!t]
    \centering
    \includegraphics[width=\linewidth]{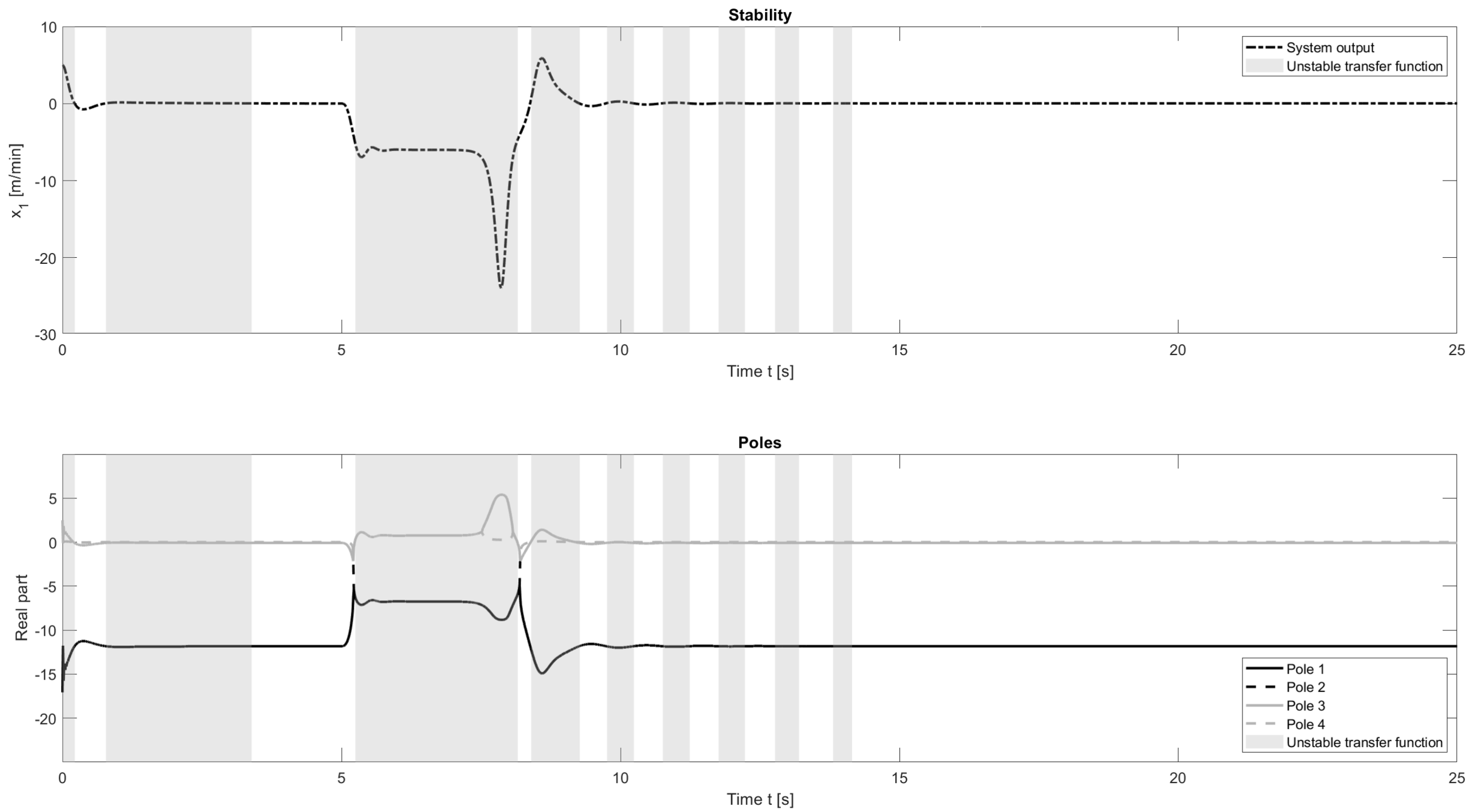}
    \caption{\textbf{Stability analysis for the chaotic thermal convection loop with disturbance.} The dashed line shows the systems output, when controlled by the neural PID controller. The closed loop transfer function is not guaranteed to be stable within the grey areas, despite the algorithm stabilizing the system. As the system approaches its steady state, the system becomes input-output stable with the chosen PID parameters. The second subplot shows the real values for all four poles. The system is only stable (white background colour) if all poles have real values smaller than zero.}
    \label{fig:stability}
\end{figure*}

\textbf{Investigating the solutions stability:}
Despite the experimental evidence that suggests the enormous benefit of using neural networks to adapt PID parameters online, this approach is not yet used on real-world system. This is due to the black box character of neural networks and the stringent safety requirements for control processes. One of the most important safety requirements of a closed loop control approach is input-output stability. It describes whether the system output is bounded for all bounded inputs. A system can be evaluated for stability by analysing the closed loop transfer function, i.e. the relation between the system output to its input.

For a bounded-input bounded-output stability analysis, the closed loop transfer function is computed as
\begin{equation}
    \frac{V(s)}{V^*(s)} = \frac{G(s)H(s)}{1 + G(s)H(s),}
\end{equation}
where $V(s)$ is the system output, $V^*(s)$ is the setpoint, G(s) is the system transfer function and H(s) is the controller transfer function in the Laplace domain. Following the Nyquist criterion, the system is stable if all poles are in the left half of the left half plane, i.e. their real values are smaller than zero.
In the example of the chaotic thermal convection loop, the systems transfer function $G(s)$, linearized around a steady state $x_1 = x_2 = x_3 = 0$, does not change over time and only has to be computed once. The controller transfer function $H(s)$ is dependent on the PID parameters and therefore changes at every time the neural network adjusts these parameters. To make sense of these changes and interpret them from a stability perspective, the controller transfer function therefore needs to be computed at every time step. This varies from the traditional stability analysis, which is computed once under the assumption of non-changing PID parameters.
Together, these transfer functions express whether the closed-loop solution is stable or unstable.

As an important contribution, we therefore perform an online analysis of the input-output stability for the controller. This analysis can be seen in Figure \ref{fig:stability}. The Figure shows the systems output and the stability, with respect to its linearized steady state over the experiment and the real values for all four poles of the transfer function. The closed loop transfer function is not stable in the beginning, during settling and after the disturbance. This can be expected, as the system is far away from its steady state for which stability is evaluated. However, as the systems output gets close to the set point, i.e. the steady state, the closed loop transfer function becomes stable. 
Knowing about the relationship between chosen PID parameters and stability allows to include this knowledge into the training. 
A potential way to include this information would be to include the poles as a regularization term during training in order to force the system towards an input-output stable behaviour. 
Furthermore, the input-output stability evaluation is an important insight for control engineers and makes the neural PID controllers understandable for humans, thus emphasising its applicability for safety critical systems.

\section{Future work}
This paper presents a first step towards accommodating the needs of control engineers when integrating machine learning algorithms into existing control architectures. While we identified a way to relate the parameters applied by the neural network back to input-output stability, this new information was not yet leveraged during the training procedure. It is a natural extension of this paper to use the newly found information about stability as a regularizer when training the neural network to ensure only input-output stable PID parameters. 
Beyond extensions of the exact framework used in this paper, we have provided an effective demonstration of the more general idea of leveraging machine learning algorithms to enhance existing control methods.

While other examples exist in the literature (e.g., the combination of neural networks\cite{liu2020time} or fuzzy logic systems \cite{liang2020event} with backstepping), many other combinations of different machine learning approaches with other control algorithms have the potential to provide good results. For example, the machine learning methods from reinforcement learning might be well-applied to linear quadratic regulators while maintaining interpretability.

\section{Conclusion}
In this paper, we conduct an extensive and rigorous investigation into the use of general dynamic neural networks for online PID parameter adaption. We perform experiments on four different systems, with and without sensor noise as well as with and without disturbance, resulting in $16$ experiments in total. These scenarios cover the most important challenges in control engineering. This study is therefore unique in its extensiveness, as previous papers only used one type of benchmark system. The neural-network-based approach outperforms a standard PID controller in $15$ of $16$ scenarios and outperforms a model-based controller in $13$ of $16$ scenarios. 

These results showcase the potential of extending existing systems by machine learning in general and neural networks in particular. Furthermore, we keep the neural network design and integration simple to allow for easy adoption of our technique. With an appropriate implementation as a library, our technique could be used without extensive knowledge of either control engineering or neural networks.
To the best of our knowledge, this is the first investigation that uses general dynamic neural networks, extending the state of the art for using neural networks to tune PID parameters. We perform a detailed analysis for one representative scenario, highlighting the superior control performance of our approach over both the traditional PID controller and model-based backstepping. It is worth noticing that while training data was gathered from simple differential equations, the results indicate significantly increased resilience towards noise and unforeseen disturbances.

Although the significant potential of neural networks for PID parameter tuning is known \cite{de2018discrete,zhang_2006}, this technique has not been used in real-world applications to date. 
As the functioning of a neural network in this setting is not understood, control engineers refrain from using them. In a first attempt to solve this problem, we perform an input-output stability analysis to interpret how neural networks function within the suggested framework. 
Tying the neural network outputs back to stability makes this neural-network-based approach understandable to humans. We therefore address a key issue when applying machine learning algorithms for control problems: the interpretability of and subsequently the trust in the machine learning solution. This work is thus an important step to increase the acceptance of machine learning based approaches for real-world systems and an important step towards safe and interpretable applied artificial intelligence.

\section*{Acknowledgements}
\footnotesize
The authors also thank Omid Namaki, Matthew Schlegel, Ahmed Shehata, Brian Tanner, and Nadia M. Ady for suggestions and helpful discussions.

\end{document}